# A Multi-input Multi-output Transformer-based Hybrid Neural Network for Multi-class Privacy Disclosure Detection

A K M Nuhil Mehdy[1] and Hoda Mehrpouyan[1]

[1]Department of Computer Science, Boise State University, Idaho, USA
akmnuhilmehdy@u.boisestate.edu, hodamehrpouyan@boisestate.edu

## Abstract

*The concern regarding users' data privacy has risen to its highest level due to the massive increase in communication platforms, social networking sites, and greater users' participation in online public discourse. An increasing number of people exchange private information via emails, text messages, and social media without being aware of the risks and implications. Researchers in the field of Natural Language Processing (NLP) have concentrated on creating tools and strategies to identify, categorize, and sanitize private information in text data since a substantial amount of data is exchanged in textual form. However, most of the detection methods solely rely on the existence of pre-identified keywords in the text and disregard the inference of underlying meaning of the utterance in a specific context. Hence, in some situations these tools and algorithms fail to detect disclosure, or the produced results are miss classified. In this paper, we propose a multi-input, multi-output hybrid neural network which utilizes transfer-learning, linguistics, and metadata to learn the hidden patterns. Our goal is to better classify disclosure/non-disclosure content in terms of the context of situation. We trained and evaluated our model on a human-annotated ground truth dataset, containing a total of 5,400 tweets. The results show that the proposed model was able to identify privacy disclosure through tweets with an accuracy of 77.4% while classifying the information type of those tweets with an impressive accuracy of 99%, by jointly learning for two separate tasks.*

## Keywords

*Feature Engineering, Neural Networks, Transfer Learning, Natural Language Processing, Privacy*

## 1. Introduction

Over the years with the increase in accessibility of internet and growth of communication platforms and social networking sites, user's concern about their privacy has also increased [31,36, 52]. In order to provide usable tools and algorithms for users to manage the disclosure of their private information, many research has been carried out [37]. Mostly focused on understanding how users are sharing their private information through emails, text messages, and social media platforms and providing them with a clear picture of privacy threats and consequences of information sharing activities [11,34].

Research in these areas is especially important, since the aggregated amount of personal information that an individual shares could be exploited by the modern AI (artificial intelligence) techniques to gain meaningful insights on their private information which could lead to serious privacy violations [20]. Wang at. al argues that user-specific targeted attacks are becoming more common by exploiting the victim's private information [49]. Hence, the need to design and develop efficient tools and techniques to protect individual's privacy have resulted in researchers focusing on understanding the individual's motive to disclose private information [23,32,53].

Researchers in the field of Natural Language Processing (NLP) have concentrated on creating tools and strategies to identify, categorize, and sanitize private information in text data since a substantial amount of data is exchanged in textual form [2,7,42]. A usable privacy-disclosure detection tool is dependent on the understanding of what constitutes as private information and what defines a disclosure for an individual user. Different information is considered as private or sensitive across different domains of human lifestyle [8]. Researchers have also intended to classify someone's private information into two main categories: objective (i.e., factual information such as age, sex, marital status, health condition, financial situation) and subjective (i.e., internal states of an individual such as interests, opinions, feelings) [45]. As per the scope of this paper, we define *privacy disclosure* as an occurrence when a piece of text, which is usually a statement/expression from an author, contains someone's private information/situation. In other words, we focus mostly on the objective disclosure where users explicitly reveal someone's privacy. We consider three types of information disclosure in this research work: health condition, financial situation, or relationship issues.

For example, a disclosure occurs when a user tweets about his/her economic situation, i.e. the financial crisis he/she is going through, investment details, etc. Another example of disclosure could be when a patient tweets about his/her own physical/mental health condition, diagnosis results, medication/drug he/she is taking, etc. The intuition is similar for the Tweets that are about relationship issues. Likewise, we define non-disclosure as an event when a piece of text is not disclosing someone's health condition, financial situation, or relationship issues. Examples of non-disclosure information sharing activities are: when an activist tweets about the national/global financial crisis, observations about the stock market, tips and tricks for the new investors, etc. Another example of non-disclosure could be when a doctor tweets about a disease, its symptoms, health care advice, etc. Therefore, a usable privacy disclosure tool is required to differentiate between public/private information and overcome the difficulties associated with the natural language processing of context-based textual data.

As part of this efforts, a wide range of proposed methodologies such as dictionary utilization, information theory, statistical model, machine learning, and deep learning have shown promising results in identifying privacy disclosure in text data [7,10,18]. However, most of the methods are based on the fact that they solely rely on the existence of keywords/terms/phrases and disregard meaning inference from the text. We observed through our experimentation that these limitations, in some cases, result in miss classification. This is because only the existence of sensitive keywords in a piece of text does not always result in user's privacy disclosure.

For example, in Figure 1, the text from the box 1 is revealing someone's health information (i.e., the patient might have cancer) and the text from the box 2 is just an article about a state of the

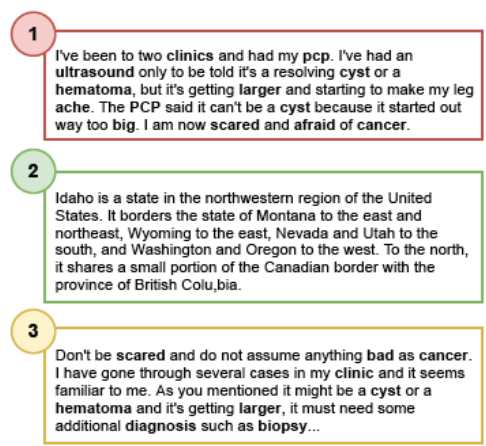

**Figure 1: Example of disclosure post (1), non-disclosure post (2), and highly similar to disclosure but actually a non- disclosure (3) [34].**

United States representing a public ambience. It is relatively an easy task in NLP to distinguish these two piece of texts based on the traditional techniques such as keyword spotting, bag-of-words model, rule based approach, etc. [5]. Now, the text from the box 3 contains similar keywords and sentence structure as the patient's post (box 1). However, this piece of text is not actually revealing the health information of the author. In other words, the doctor does not have cancer rather just a comment about cancerous disease. Therefore, it is quite challenging to distinguish the content of box 1 from the box 3, without taking into the consideration the underlying meaning and hidden patterns.

## 1.1 Contributions of the Work

We propose a novel and hybrid multi-input multi-output neural network based model that overcomes the NLP challenges by precisely identifying privacy disclosures through tweets by combining knowledge from pre-trained language model, semantic analysis, linguistics, and the use of metadata. The multi-input, multi-output model is able to identify both the information type (health, finance, relationship) of the tweets and the disclosure occurrence by jointly learning for two separate tasks. We also trained and evaluated our model on a human annotated ground truth dataset that contains a total of 5,400 tweets from anonymous users. Thus, our model could be implemented in the practical and usable fields of data privacy, information security, natural language processing, etc. A few of the notable contributions of this paper includes:

- Presenting a multi-input, multi-output hybrid neural network that utilizes pre-trained language model, and still make use of traditional linguistics and structured metadata.
- Evaluating its multi-output capability that jointly learns for solving two separate NLP tasks while utilizing a pre-trained language model.
- Sharing the model performance on a ground truth dataset for benchmarking.

The rest of the paper is organized as follows: section 2 contains the background and reviews on the related research works and a few of their limitations we observed. Section 3 describes about the dataset used in this paper along with the detail of the data labelling strategy. The methodology, data pre-processing, and feature engineering techniques are described in detail in section 4. The detail of the deep neural network architecture is presented in section 5 following the experiments in 6. Lastly, section 7 represents the experimental results following the conclusion.

## 2. BACKGROUND AND RELATED WORK

In this section, we review the state-of-the-art natural language processing research that are focused on privacy disclosure detection [3,8,22,33]. Traditional research in this field has mostly relied on lexicon-based techniques to automate the content analysis of privacy-related information by leveraging linguistic resources such as privacy dictionaries. Existing automated content-analysis tool such as LIWC[1] is used with a specific sets of privacy dictionaries. Vasalou et al. suggest such a method that utilizes a dictionary of individual keywords or phrases which are previously assigned to one or more privacy domains [47]. To create the dictionary, they sample from a wide range of privacy domains such as self-reported privacy violations, health records, social network sites, children's use of the Internet, etc. However, their technique solely relies on the predetermined sensitive keywords/terms which classify both a medical article (public) and someone's medical condition (private) as a private document. Similar approach was taken by Chakaravarthy et. al. for a document sanitization task, where they represent a scheme that detects sensitive information using a database of entities [7]. The database contains different entities i.e. persons, organizations, products, diseases, etc. Each entity is also associated with a set of sensitive terms e.g. name, address, age, birth date, etc. Thus, a set of terms is considered as the context of the entity. For example, the context of a person become his/her age, birth date, name, etc.

---

[1] Linguistic Inquiry and Word Count

Researchers from the area of information theory leverage large corpus of words along with computational linguistics to identify sensitive information in text documents [42]. Information theory provides the necessary formula for calculating the sensitivity score, otherwise known as IC (Information Content) score of every term, based on the amount of information it contributes to a corpus. For example, in a database of employee, a term such as *handicapped* carries more information than the common terms such as job, manager, desk, office, etc. All such terms that exceed a threshold score β are considered as sensitive. One of the advantages of this technique is that a finite collection of named entities is not required for the disclosure detection to be successful. However, this approach suffers from the same limitation as the previous line of work. In other words, it does not consider any semantic information other than merely relying on the appearance of sensitive keywords. Other popular techniques such as Named Entity Recognition (NER), also known as entity chunking, entity identification, or entity extraction have also been used by many researchers to identify and classify private information in text documents [2]. This line of research is based on the sub-task of information extraction technique that aims to identify named entities (medical codes, time expressions, quantities, monetary values, etc) and classify them into predefined categories in an unstructured text. Modern NER systems use linguistic grammar-based techniques, statistical models, machine learning, etc. Regardless of the underlying method, the NER based disclosure detection techniques also lack the capability of properly inferring the meaning from a text that could disclose someone's private information if a specific named entity is not detected (see examples 3 in Table 1).

Machine learning based techniques such as association rule mining [10], support vector machines (SVM), random forests [45], boosted Naive Bayes, AdaBoost, latent Dirichlet allocation (LDA), etc. have also been used to tackle similar tasks. Hart et al. used a novel training strategy on top of SVM to classify text documents as either sensitive or non-sensitive [18]. Caliskan et al. proposed a method for detecting whether or not a given text contains private information by combining topic modelling, named entity recognition, privacy ontology, sentiment analysis, and text normalization technique [6]. A combination of linguistic operations and machine learning is proposed by Razavi et. al. to detect health information disclosure [38]. They first compile a list of keywords related to a person's health information, and then apply keyword combinatorial web search. Alongside, they implement a machine learning layer to detect and learn any possible latent semantic patterns in the annotated dataset. Mao et al. studied privacy leaks on Twitter by automatically detecting vacation plans, tweeting under influence of alcohol, and revealing medical conditions [33] As the classifier model, they implemented two machine learning algorithms; Naive Bayes and SVM based on the TF-IDF (Term Frequency Inverse Document Frequency) feature space. Their main research goal was to analyse and characterize the tweets in terms of who leaks the information and how. Therefore, in the paper, the focus was less on the architecture and performance of the disclosure detection model.

Bak et. al. applied a modified LDA based topic modelling technique for semi-supervised classification of Twitter conversations that disclose private information [4]. This technique is also based on the distributions of terms/keywords across documents and corpus, which again does not consider word meaning inference. Most of the above-mentioned approaches have drawbacks since they rely exclusively on the presence of keywords and ignore word meaning inference from the text. We observe through our experimentation that, these limitations, in some cases, result in miss classification. This is because, existence/lack of sensitive terms/keywords in a piece of text does not always result in disclosure/non-disclosure of private information (see examples 3,4,5 in Table 1).

In order to overcome these limitations, recent research works from the area of NLP and privacy have considered utilizing semantic meaning along with lexical and syntactic analysis, while designing and developing deep learning based models [12,34,35,46]. Accordingly, there have been a significant progress in the area of language modelling through training complex models on enormous amounts of unlabelled data [14,48]. All the tailored solutions are being outperformed by this generic models. Most importantly, the utilization of transfer learning and pre-trained model have shed light into this area of research. Dadu et. al. proposed a predictive ensemble model by exploiting the fine-tuned contextualized word embedding, RoBERTa (Robustly Optimized BERT Approach) and ALBERT (A Lite version of BERT). The authors generated a small, labelled dataset, containing Reddit comments from casual and confessional conversations. Through the ensemble implementation they achieved 3% increment in the F1-score from the baseline model. Therefore, after considering the importance of transfer-learning and also taking into account the significance of linguistic features, we propose a multi-input hybrid neural network which utilize both transfer-learning and linguistics along with the metadata from the input text. The multi-output model is also able to classify both the information type of the input text and the disclosure occurrence by jointly learning for two separate tasks. We next describe the proposed framework.

## 3. DATASET

The deep learning based methodology proposed in this paper consists of a supervised neural network model that requires labelled data to learn the patterns of the disclosure and non-disclosure texts. There might be several reasons why no dataset is available for this purpose in literature, i.e., the restricted access policies of such data sources (e.g., emails, SMS, chat records), lack of privacy preserving research strategies, the complexity associated with the data labelling technique, etc. Therefore, we collected, and human annotated a ground truth dataset that contains human expressions, comprised of multiple English sentences, through which their privacy might have been disclosed. The following two sections detail our data collection and data labeling steps.

### 3.1 Data Collection

In order to collect diverse and user-centric data from different domains, we use the online platform, Twitter. People tend to prefer this platform to share their personal opinions, perceptions, issues, and observations through tweets which are comprised of a few sentences, hashtags, and emojis. We utilized Twitter search API [44] for mining the required dataset following a set of cleaning and labelling processes. We limited the data collection to those tweets that are written in English language and from anywhere in the world. This allows us to collect a generalized set of data written in different styles. The dates for crawling the tweets are randomly chosen for better sampling. Most importantly we filtered out the tweets based on a set of criteria such as i) tweets that contain any links, ii) retweets iii) replies to the tweets iv) tweets that are from verified accounts v) tweets that are posted by bots.

A total of 45,000 tweets is collected from three different privacy domains i) health, ii) finance, iii) relationship. The advanced search query strategies offered by the Twitter API [44] allowed us to properly identify and collect the tweets from these three categories. From these sets of tweets, we sampled a set of 6,000 random tweets based on the stratification on these three information types, selecting 2000 tweets from each category. This smaller subset of dataset is then used for human annotation and model training. In addition, we maintained the anonymity of the tweets by removing all the metadata excepts the tweet's date-time, tweet texts, and device-type used to post these tweets. Therefore, usernames, handles, permalinks, or tweet id remained hidden from the human annotators as an ethical consideration. We also meet the Twitter Developer Agreement and Policy[2].

### 3.2 Data Labelling

In each of the collected tweets, people tend to share their personal issues, opinions, perceptions, and advice, etc. It is observed that the authors intentionally or unintentionally disclose their own or someone else's private information such as health condition, financial situation, or relationship issues through their tweets. Some examples of such privacy disclosure and non-disclosure tweets can be found in Table 1 which are randomly sampled from the 6K dataset.

---

[2] "You may use the Twitter API and Twitter Content to measure and analyze topics like spam, abuse, or other platform health-related topics for non-commercial research purposes by conducting only non-commercial research on this dataset."

Table 1: Example of disclosure and non-disclosure tweets
(Samples are taken from the set of 6,000 tweets).

| No | Text | Information Type | Is a Disclosure? |
|---|---|---|---|
| 1 | Ran into two 'mean girl' ex friends today. They're still mean. I was having a bad mental health day too. But I'm choosing to look on it as a lesson that I was right to cut them off. I was having doubts about one of them. Not now. | Health | Yes |
| 2 | stop calling me a homewrecker I'm simply breaking up a relationship for my own personal gain RANBOO HELLO | Relationship | Yes |
| 3 | We all 7311 Candidates who passed Beltron Deo 2019 2020 exam want joining because our financial condition is so poor and all are workless. | Finance | Yes |
| 4 | Financial abuse is so scary amp it's very common. It's why I always discourage women from being transparent about their finances (he doesn't need to know about all your money) or merging finances with a man and not having her own private accounts. | Finance | No |
| 5 | Being self aware is sexy. Taking your mental health serious is sexy. Loving yourself sexy. Pretty face and body fades eventually but your mind will always keep developing and expanding. | Health | No |
| 6 | Shout out the teachers who talked about their divorce and personal problems and just passed us instead of teaching | Relationship | No |

We recruited human annotators from Amazon Mechanical Turk[3], an online crowd-sourcing marketplace to label all of the tweets as either disclosure or non-disclosure. The detailed instructions along with a set of good and bad examples of labelling was provided to assist the annotators understand the task correctly. We specifically guided them to follow the definitions of disclosure and non-disclosure, provided in section 2. We limited the selected annotators to USA with a good reputation (i.e., at least 95% HIT[4] approval rate) and those who are at least 18 years old. Each annotator was paid $0.05 per tweet based on our pilot trials indicating workers could label each tweet within 30 seconds. It is worth mentioning that only the binary labelling of disclosure/non-disclosure was completed by the human annotators. They were not asked to label the information types, since we already assigned these labels as a bi-product while crawling the tweets using the advanced search query API of Twitter. Most importantly, we employed 3 human annotators per tweet to decide whether or not that post is a privacy disclosure. This enabled us to select the most voted label for the tweet as the ground truth.

### 3.3 Data Augmentation

We discovered a moderate level of data imbalance after annotating the dataset. A total of 807 tweets out of 2000 from the health category and 769 tweets out of 2000 from the finance category were labelled as disclosure class whereas 799 tweets out of 2000 from the relationship category were labelled as non-disclosure. Therefore, we performed a data augmentation step to make the dataset balanced. First, we randomly sampled the candidate tweets to be augmented from each category. We sampled 93 disclosure tweets from the health category, 131 disclosure tweets from the finance category, and 101 non-disclosure tweets from the relationship category. Then we applied *domain-specific paraphrasing and synonym replacement* technique on these tweets as our augmentation strategy. This simple yet effective approach of augmenting text data has been recommended by the researchers and proved to be useful for getting generalized text data [50]. After the augmentation, we got 900 (800+93) disclosure tweets for the health category, 900 (769+131) disclosure tweets from the finance category, and 900 (799+101) non-disclosure tweets for the relationship category. On the other hand, we re-sampled 900 non-disclosure tweets from the health category, 900 non-disclosure tweets from the finance category, and 900 disclosure tweets from the relationship category. This resulted in a balanced dataset of 5,400 tweets where each of health, finance, and relationship categories contained 900 disclosure and 900 non-disclosure tweets (Table 2).

---

[3] A crowd sourcing website for businesses and researchers to hire remotely located "crowdworkers" to perform on-demand tasks such as survey, data labelling, etc.
[4] Human Intelligence Task

**Table 2: Final dataset (balanced) for model training.**

| Info Type | # of Disclosure Tweets | # of Non-disclosure Tweets | Total |
|---|---|---|---|
| Health | 900 | 900 | 1800 |
| Finance | 900 | 900 | 1800 |
| Relationship | 900 | 900 | 1800 |
| Total | 2700 | 2700 | 5400 |

## 4. METHODOLOGY

The neural network based model proposed in this paper adopts a transformer based pre-trained model called BERT (Bidirectional Encoder Representations from Transformers). We use this state-of-the-art pre-trained model to develop our custom multi-input multi-output model because: i) it supports fine-tuning for custom NLP tasks (transfer learning), ii) it is trained on a huge corpus of unlabelled texts (3,300 millions of words), ii) contains millions of parameters (110M), iv) supports parallelization for hardware acceleration, etc. The following subsections further detail on this component along with the data pre-processing and feature-engineering steps.

### 4.1 Data Pre-processing

As depicted in Table 1 both disclosure and non-disclosure tweets could contain similar keywords, sentence structure, and other syntactic constructs. This makes the classification problem particularly challenging, because we cannot simply rely on the lexical items and obvious keywords in the text, like bag-of-word models. Rather, we are required to discover the hidden patterns and infer author's intentions that are embodied in the text, and to encode the underlying meaning expressed in the text to better classify the disclosure/non-disclosure activities. Therefore, unlike the traditional approaches that are mostly based on the bag-of-words technique, we kept the punctuation and stop words in the text to preserve the syntactic structure. We use NLP Toolkit [19] to clean the tweets in a customized way that ignores noisy and redundant tokens such as *",,"*, *";--"*, *"!!!"*, *":-)"* and preserves the non-redundant ones such as *","*, *";"*, *":"*, *"."*, *"he"*, *"the"*, *"in"* etc. This is in contrast to the traditional approach of text analysis that is based on removing all the punctuation. It is important to note that we also removed Twitter specific tokens such as *@, #* and non-unicode special characters which might have been added by the users' device.

### 4.2 Feature Engineering

We performed feature engineering on the dataset to produce four new features which then were fed into the neural network through its multiple input channels. Based on the Dependency Parse (DP) tree information of the texts, the underlying syntactic relationship of the data was generated. Additional features, i.e. date and time of the tweets, and the type of device that was used to post the tweets are also fed into the network as a meta data. Below we explain these new synthetic features in more details.

### 4.2.1 Syntactic Structure

Certain formal properties of the language such as dependency parse tree information are known as a ''purely stylistic'' by the theoretical linguistics [4]. In other words, two English sentences might have different syntactic forms but still express the similar meaning or vice versa [15]. For example, (*I suffered a lot in last few days*) with the DP structure *nsubj ROOT det dobj prep amod amod pobj* could be semantically equivalent to another sentence (*In last few days I suffered a lot*) having the structure *prep amod amod pobj nsubj ROOT det npadvmod*, though they are syntactically different. Figure 2 depicts the DP information of an example sentence where the DP tags are shown on the edges.

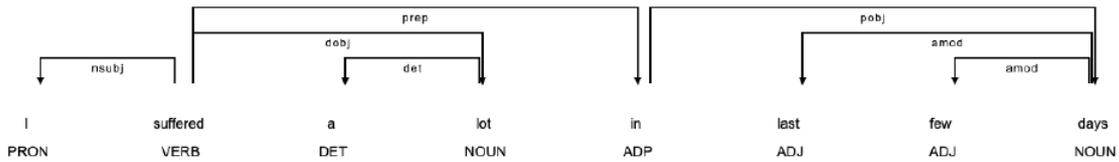

**Figure 2: Dependency Parse Tree Information of a Sentence.**

Along with the parts of the speech tags, these types of representation of the language features enable the deep-learning based models to learn about the sequential patterns of the sentence constructs along with the arrangements of the word token themselves [35]. This helps the model. Hence, we used a natural language toolkit [19] to extract the information to enrich the feature space of the dataset.

### 4.3 Transfer Learning and Fine Tuning

Most of the NLP tasks such as text classification, machine translation, text generation, language modelling, etc. are considered as sequence modelling tasks. Typical machine learning models such as bag-of-words, term-frequency inverse document-frequency, and multi-layer perceptron are not able to capture the sequential information presented in the text. Therefore, to capture this important piece of information, researchers have introduced techniques such as recurrent neural network (RNN) and long short-term memory-based network. However, these types of neural networks introduce new issues in terms of performance and efficiency. For the reason that both RNN and LSTM based neural network takes one input (token in case of text sequence) at a time, they could not be parallelized. This makes the training operation, time consuming, specially while handling a large dataset.

This was the case until 2018 when Google introduced the transformer model which turned out to be ground-breaking [48]. It is mainly an attention mechanism for learning contextual relations between words in text (Figure 3).

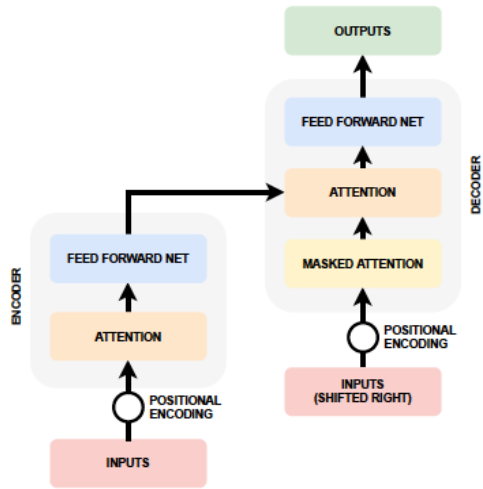

Figure 3: Simplified View of the Transform Architecture [48]

It also introduced an architecture that supports parallelization and make use of unlabelled text data for training. In the following year, BERT has been introduced which makes use of the transformer architecture. It is a new language representation model published by the researchers from the Google AI Language team in 2018 [14]. Since then all the tailored solutions to various NLP tasks are being outperformed by this generic transformer based model. Most importantly, BERT supports transfer-learning which allows us to develop domain specific custom NLP models while utilizing the power of transformer based pre-trained models. Transfer learning is pre-training a neural network model on an informed task and then using the trained network as the basis of a new purpose-specific model, otherwise known as fine-tuning [43]. Researchers from the area of computer vision have already shown the significance of this technique [17], and in recent years, they have been showing how a similar technique could be useful in natural language tasks as well [40]. Figure 4 depicts an abstract view of BERT's pre-training and fine-tuning Procedures.

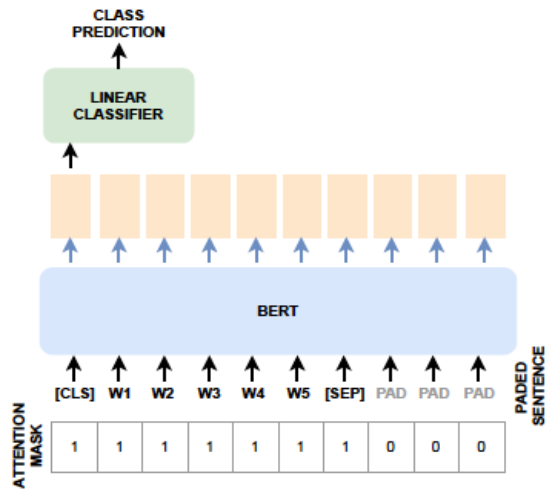

Figure 4: Simplified View of BERT Fine-tuning Procedures [14]

Therefore, we adopt transfer-learning technique to design and develop our hybrid multi-input multi-output neural network which not only fine-tune a pre-trained model but also make use of the linguistic pattern-learning and metadata utilization. There are different ways of fine tuning a model: i) the entire architecture could be further trained on a new dataset which allows the model to update its pre-trained weights ii) retraining only the higher layers while keeping the weights of initial layers of the model frozen iii) keeping the all the layers of the model frozen, and add one or more new neural network layers of our own, where only the weights of the new layers will be updated during the training phase. In this paper, we utilize the last technique where we import the pre-trained BERT model as a neural network layer into our custom neural network architecture. This acts as one of the three main input channels of our network. The other two input channels provide additional data to the model that we detail in the following sections.

% Using transformer-based models has multiple advances including but not limited to - i) they can take the entire sequence of tokens as text input enabling the capability of training acceleration by GPUs and TPUs, ii) no need of labelled data for pre-training the model, iii) they are better for transfer learning iv) supports better model explainability. Most significantly, this pre-trained model can be fine-tuned with just one extra output layer to produce state-of-the-art models for a wide range of applications with little task-specific architectural changes. Having the ability to be fine-tuned is also advantageous because these types of models, for example BERT, consists of a huge number of parameters (100M - 300M). Therefore, training such a model from the scratch on a relatively smaller dataset may result in poor performance (e.g., over-fitting or under-fitting).

## 5. NEURAL NETWORK ARCHITECTURE

The architecture of the proposed neural network is divided into three main segments: i) pre-trained BERT model ii) implementation of the linguistic features iii) integration of structured metadata. The output of this model consists of two different branches: i) multi-class classification of information types ii) binary classification of disclosure/non-disclosure information sharing transactions. Figure 5 depicts the architecture of the proposed multi-input, multi-output hybrid neural network. In the following subsections, we describe each component of the model in detail.

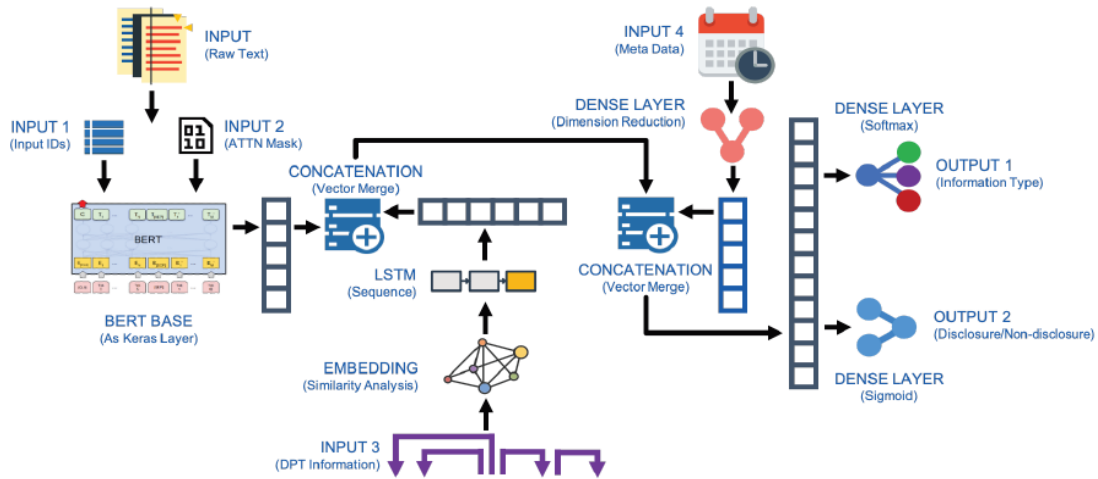

Figure 5: Architecture of the Proposed Model

### 5.1 Leveraging BERT

BERT model has two inputs: first from the word tokens, and second from the segment layer following their embedding layers. BERT has a vocabulary of 30,000 distinct tokens comprised of complete English words and word piece components (e.g., embedding for both *play* and *##ing* to work with *playing*). These tokens are associated with an initial embedding space known as WordPiece embedding. The two inputs are added and summed over a third embedding known as position embedding, followed by the dropout layers and layer normalization. The resulting BERT model contains 12 multi-headed self-attention layers (encoders) which are identical to each other. BERT is trained on two NLP tasks: i) the Next Sentence Prediction (NSP), ii) Masked Language Modelling (MLM). These two tasks are informally called fake tasks. In other words, when the pre-training of BERT happens, the model learns the language patterns while solving for these two given tasks. In the end, the trained model is saved and used for further fine-tuning to solve specific NLP tasks, like one in this paper (disclosure and information type classification). Please refer to the original research paper for a detailed implementation of the BERT's architecture [14].

### 5.2 Inputs to the Proposed Network

As mentioned previously, we import the pre-trained BERT model as a layer into the proposed neural network. Token-ids and attention-masks are fed into the model through two input channels of the BERT layer. Token ids are the integer encoded values for each of the tokens of the input text. Attention masks are supporting vectors that enable BERT differentiate between the actual and padding tokens. We add a dropout layer after the BERT main layer as suggested by the literature [21]. In addition, a separate input channel was added to the proposed neural network through which we fed the dependency parse tree information of the same input text. This input path has its own embedding layer which gets learned during the training process. Then we added an LSTM layer that learns the sequential information of the dependency tags. Output of this LSTM layers are then concatenated with the output of the dropout layer. At this stage, we employ the metadata to the neural network through another input channel. This input takes the day of the week, hour of the day, and device type information associated with each input text. A dense layer is added to reduce the dimensionality caused by the encoding of these categorical features. This dense layer uses rectified linear unit as its activation function. Finally, we concatenated the output of this input channel with the output of the previous concatenation operation (BERT's output + DP output).

### 5.3 Outputs from the Proposed Network

Since we aim to solve two parallel tasks through a single neural network model, there are two separate output layers in the proposed model. In one output layer we add three neurons that result a probability distribution of the information type variable. The predicted probabilities of an input text being any of the three classes: health, finance, relationship is distributed among these three neurons. The neuron with the highest probability wins and shows the information type of the input text. The other output layer is comprised of a single neuron which calculate the probability of the input text being either disclosure or non-disclosure. In other words, the model jointly optimizes for a multi-class classification task and a binary-class classification task. Therefore, we employ different loss functions for these two separate out layers. The multi-class prediction layer uses categorical cross entropy, and the binary class prediction layer uses binary cross entropy with accuracy as the evaluation metrics.

## 6. EXPERIMENTS

In this section we describe the implementation detail of the proposed neural network architecture along with the tools we used. We also talk about the optimizer, loss functions, metrics, and a set of hyper-parameters in this section.

### 6.1 Tools and Libraries

We utilize the Huggingface's Transformers package which is an open source natural language processing library developed in Python programming language [51]. This library lets developers import a wide range (32+ pre-trained models in 100+ languages) of transformer-based pre-trained models such as BERT, ALBERT, XLnet, GPT-2, etc. It is also very easy to switch between different transformer based models through Huggingface Transformers. Most importantly, it supports interoperability between PyTorch, TensorFlow, and other deep learning libraries. We use Tensorflow that comes with Keras pre-built to architect the multi-input multi-output neural network [1]. More specifically, we use the the Keras functional API to create the neural network architecture [26].

We make use of the *TFBertModel* module from the Transformers package which is an interface to the Tensorflow library. We import the pre-trained BERT model called *bert-base-uncased* using this module. This is a pre-trained model on English language, and it is uncased meaning it does not make a difference between the words playing and Playing [13]. This specific base model consists of 110 million parameters. The main layer of this pre-trained model is imported as a keras layer into our custom architecture following a dropout layer. In the other input channel of our model, a LSTM layer with *tanh* activation function is used over the dependency parse tree information by utilizing the keras *LSTM* layer [27]. Before this layer, we use the keras *Embedding* layer to learn the embedding of these dependency tags in a 16-dimensional vector space [25]. The Keras *concatenate* method then takes the output from this LSTM layer and the dropout layer from BERT to merge them into a single vector. The final input into our custom neural network makes use of a keras *Dense* layer [24], and its output is also gets concatenated with the other branch before going through the final output layers.

For text pre-processing, we applied Spacy [19] to derive the dependency parse tree information of each tweet. Spacy provides dependency parser, trainable models, tokenizer, noun chunk separator, etc. in a single toolkit. It offers the fastest syntactic parser in the world and its accuracy is within 1% of the best available natural language toolkit [9]. To perform the data augmentation step, we used another Spacy based library called spaCy WordNet [39]. It is a custom component for using WordNet and WordNet domains with spaCy which allows users to get synsets for a processed token filtering by domain. Text encoding and padding for these tag based sequences are done using Keras text to sequence and padding methods respectively [28]. To tokenize, pad, and prepare the raw texts for the BERT side input, we utilize the *BertTokenizerFast* that comes with the Transformer package. This tokenizer converts the raw texts into BERT compatible format such as adding special tokens ([CLS], [SEP]), truncating longer sequences, returning token ids and attention masks, etc.

### 6.2 Optimizer, Loss, and Metrics

We use *Adam* gradient descent algorithm as the optimization method for the neural network. It is considered to be computationally efficient and has little memory requirement [30]. The separate output heads use two different logarithmic loss functions: categorical cross entropy for information type classification, and binary cross entropy for the disclosure detection. The network uses *accuracy* as the optimization metrics for both of the output heads which is evaluated by the model during training and testing.

### 6.3 Hyper-parameters

In case of fine-tuning based training, most of the hyper-parameters of the core model itself stay the same. Therefore, we also retain the hyper-parameters of BERT as it is. However, readers can refer to the BERT paper which gives specific suggestions on the hyper-parameters that require further tuning. In this section we only describe the about those hyper-parameters which we use for our custom neural network model.

First of all, we consider 55 (mode) as the maximum length of the input text sequences. Since the tweets in the dataset are of varying length, we use truncation and padding to make all the tweets have this same length. The first custom input that takes the dependency parse tree information learns an embedding space of length 16 with a vocabulary size of 47. The subsequent LSTM layers is comprised of 32 units which is the dimensionality of its output space. This layer uses *tanh* as the activation function with no *dropout*. All other parameters are kept default from the keras implementation [27]. The Keras *concatenate* method takes the output from this LSTM layer (32 dimensions) and the dropout layer from BERT (768 dimensions) to merge them into a single vector of 800 dimensions. The other custom input channel (metadata input) uses a dense layer with 32 neurons and rectified linear unit as their activation function which reduces its 149 dimensional input to 32. One of the final output layers that classify the information type uses 3 neurons with *softmax* activation. The other output that detects disclosure uses a single neuron with *sigmoid* activation. Both of these output layers use truncated normal distribution as the kernel initializers where the standard deviation is 0.02 for initializing all weight matrices. This value comes as default from the standard implementation of BERT by the Transformer library. The parameters for the Adam optimizer are chosen as follows: learning rate = *5e-04*, epsilon (a small constant for numerical stability) = *1e-08*, clipnorm (gradient norm scaling) = *1.0*. Other parameters of this optimizer are kept as default from the Tensorflow implementation it [29].

The whole dataset is split into a 90-10 ratio for training and testing respectively. For the validation purpose, we kept 20% from the training dataset while the model training process happens. Thus, 10% of the original dataset are used as test dataset which was never shown to the model. We feed the input data to the model with a batch size of 64, and let the model train for 5 epochs. We achieved the best performance from the model withing this amount of iterations. It's worth mentioning that, all the above mentioned hyper-parameters are chosen based on several trials and outcomes.

### 6.4 Computing Resources

We used Google Colaboratory [16] as the experimentation platform which provided a Nvidia Tesla T4 GPU with 16GB memory. It took 15 minutes in average to run a complete training phase given the hyper-parameters that we mentioned already. Since this platform provides virtual infrastructure and sometime shares the resources among the users, the reported time may vary.

## 7. RESULTS

The results show that, by utilizing transfer learning and pre-trained language model, a multi-input neural network based model can be trained that learns beyond simple keyword spotting and utilizes linguistic features to classify whether or not a piece of text contains a privacy disclosure with a useful degree of accuracy. Moreover, through the experimentation, it is observed that, integration of metadata to the model increases the performance noticeably (increasing the accuracy by 1.80%). Since our dataset is balanced, we report receiver operating characteristic (ROC) curve, precision[5]. and recall[6]. score, f1-score[7], confusion matrix, and accuracy[8] score for both the binary and multi-class classification task.

### 7.1 Evaluation Considerations

The classification of *information type* is not the main and only evaluating pillar of the proposed model; rather, the classification of the *disclosure vs. non-disclosure* text is the main focus of the paper. In other words, our multi-input multi-output model is designed to solve the challenging task of distinguishing highly similar texts into disclosure and non-disclosure class. The *information type* classification is a bi-product while jointly training the multi-output model. Also, since we collected the tweets from three different information domains by utilizing the Twitter API, the texts are already well-aligned with these three classes. Therefore, this is expected to achieve a higher degree of accuracy while classifying the information types. However, classifying those tweets as disclosure vs. non-disclosure becomes a crucial challenge to solve. It can also be observed from Figure 6, which depicts how various models struggle to achieve better accuracy on the binary classification task. This is expected, as we described earlier, the textual similarities between these two classes of texts. It is also evident that a binary classification task on top of highly similar texts is still a challenge [12].

---

[5] What fraction of predictions as a positive class were actually positive.
[6] What fraction of all positive samples were correctly predicted as positive.
[7] The harmonic mean (average) of the precision and recall.
[8] The fraction of the total samples that were correctly classified.

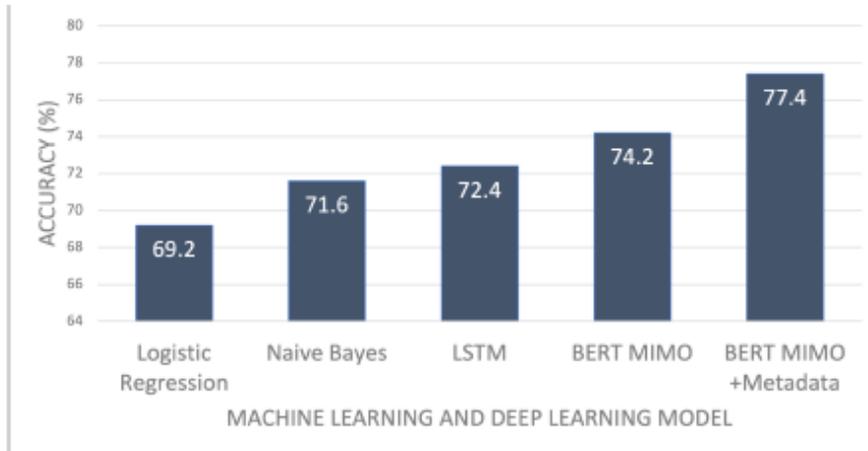

**Figure 6: Comparison among the binary classification (dis- closure vs. non-disclosure) models.**

In Table 3 and Table 4, we describe the classification report for both information type and disclosure detection respectively. As can be seen from these tables, the information type classifier achieves an impressive accuracy of 99%. The disclosure/non-disclosure classifier reaches up to 77.4% which is 8.2% more than bag-of-words and RNN based baseline models. We can also see a good recall score for the binary classifier which depicts its capability to detect most of the disclosure texts. In other words, 77% of all the disclosure texts have been identified successfully.

**Table 3: Classification report for Information Types on the Test Dataset (10% of 5400=540)**

|  | Precision | Recall | f1-score | Support |
|---|---|---|---|---|
| Health | 0.99 | 0.99 | 0.99 | 180 |
| Finance | 0.99 | 1.00 | 1.00 | 180 |
| Relationship | 1.00 | 0.99 | 0.99 | 180 |
| Accuracy |  |  | 0.99 | 540 |
| Macro Avg. | 0.99 | 0.99 | 0.99 | 540 |

**Table 4: Classification report for Disclosure/non-Disclosure on the Test Dataset (10% of 5400=540)**

|  | Precision | Recall | f1-score | Support |
|---|---|---|---|---|
| Disclosure | 0.78 | 0.76 | 0.77 | 270 |
| Non-disclosure | 0.76 | 0.79 | 0.78 | 270 |
| Accuracy |  |  | 0.77 | 540 |
| Macro Avg. | 0.77 | 0.77 | 0.77 | 540 |

Figure 7 depicts the confusion matrix for information type classification. It can be seen that, only a few miss-classifications have occurred specially when the information type of the texts was *Relationship*. Likewise, Figure 8 depicts the confusion matrix for disclosure/non-disclosure classification. In Figure 9, we show the ROC curve for information type classification, and in Figure 10 we show the ROC curve for disclosure/non-disclosure classification. The binary classifier shows an area under curve (AUC) score of 0.834. Unlike the binary class ROC curve, we render the multi-class ROC curve by using one-vs-all technique to properly represent its performance.

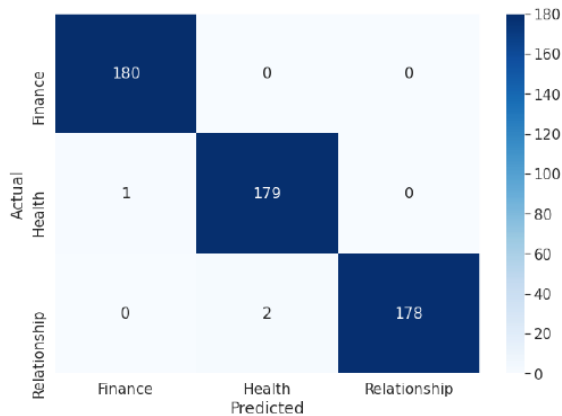
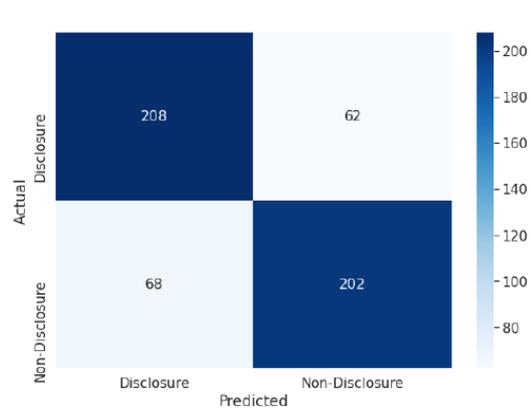

**Figure 7: Confusion matrix for information type classification**

**Figure 8: Confusion matrix for disclosure classification**

The performance of our model is not directly comparable with other similar approaches proposed in the literature because of the lack of common and shared dataset with similar properties. However, the closest and recent work of detecting self-disclosure on the #OffMyChest dataset, which contains Reddit comments, is worth comparing [12]. In their work, they achieved an accuracy of 74.12% and 74.20% on two different classes of the dataset: information disclosure and emotional disclosure respectively. Also, the precision and recall scores were 0.710, 0.551, and 0.636, 0.510 respectively. In comparison, the performance of our model is noticeably better in all the metrics.

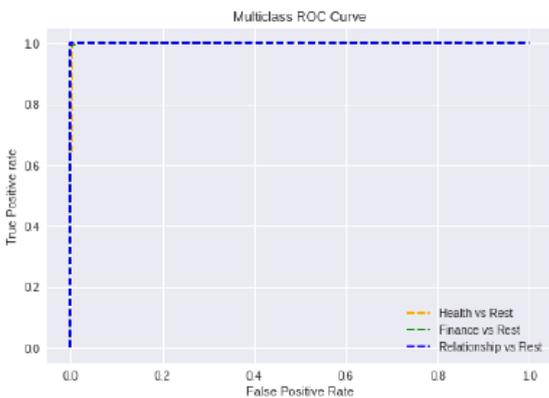
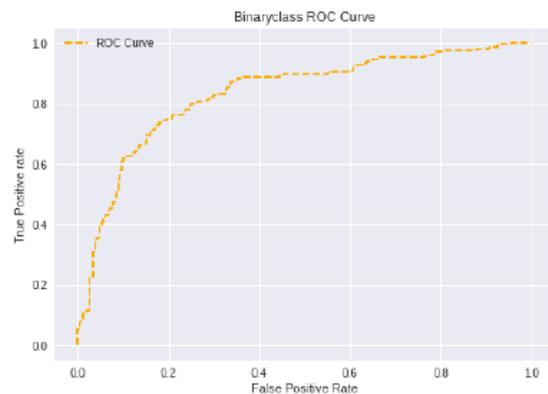

**Figure 9: ROC Curve for Information Type Classification**

**Figure 10: ROC Curve for Disclosure Classification**

## 8. CONCLUSION

In this paper we have proposed a multi-input, multi-output hybrid neural network that utilizes state-of-the-art transformer based pre-trained model called BERT along with language features and metadata to precisely detect privacy disclosure in text data. We also evaluate our model on a ground truth dataset that contains a total of 5,400 tweets from three different privacy domains: health, finance, and relationship. Unlike the traditional text classification techniques that primarily rely on keyword spotting, this model focus on underlying meaning and hidden patterns by leveraging pre-trained language model and classical linguistics. Additionally, our proposed architecture shows capability of solving two separate text classification tasks withing a single model that provides new insights which can help build practical NLP models. However, there are improvement scopes in the work presented in this paper. The learning and predictive performance of the model can be evaluated on a diverse dataset by taking samples from different data sources. Therefore, we want to collect a diverse dataset on various privacy domains in the future, using more sources such as forums, emails, text messages, and so on. In addition, performing privacy-preserving text analysis, and testing the integration of the model to the end products could also be future works. Most importantly, we plan to integrate explainability into the model for its fairness and trustworthiness.

## ACKNOWLEDGEMENTS

The authors would like to thank National Science Foundation for its support through the Computer and Information Science and Engineering (CISE) program and Research Initiation Initiative(CRII) grant number 1657774 of the Secure and Trustworthy Cyberspace (SaTC) program: A System for Privacy Management in Ubiquitous Environments.

## Authors

**A K M Nuhil Mehdy** completed his Ph.D. degree from Boise State University, USA under the Cybersecurity emphasis, in Summer 2021. Earlier, he completed his B.Sc. Eng. degree in Computer Science back in 2011 from Rajshahi University of Engineering and Technology, Bangladesh and M.S. in Computer Science in 2017 from Lamar University, USA. His research interests include Privacy and Security, related to the internet users, Industrial Control Systems, Internet of Things, and Distributed Systems. Currently he is working as a Machine Learning Engineer at Micron Technology, Inc., USA.

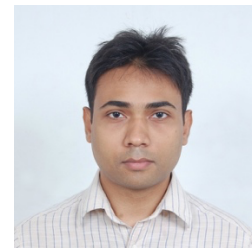

**Hoda Mehrpouyan** is an associate professor of the Computer Science department at Boise State University, USA. Dr. Mehrpouyan's research focuses on ensuring privacy, security, and robustness of mission-critical cyber-physical systems. In May 2019, she was awarded a National Science Foundation CAREER Award from the Secure and Trustworthy Cyberspace (SaTC) program. Further, she received funding from the National Security Agency (NSA), Idaho National Lab (INL), and Idaho Secretary of State. She has more than 35 peer reviewed publication.

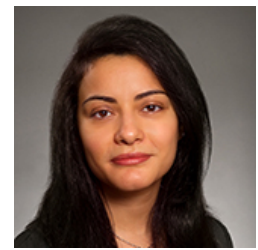